\documentclass{svproc}
\usepackage[utf8]{inputenc} % allow utf-8 input
\usepackage[T1]{fontenc}    % use 8-bit T1 fonts
\usepackage{hyperref}       % hyperlinks
\usepackage{url}            % simple URL typesetting
\usepackage{booktabs}       % professional-quality tables
\usepackage{amsfonts}       % blackboard math symbols
\usepackage{nicefrac}       % compact symbols for 1/2, etc.
\usepackage{microtype}      % microtypography
\usepackage{graphicx}
\usepackage{subfigure}

\begin{document}
\mainmatter              % start of a contribution

\title{Ursa: A Neural Network for Unordered Point Clouds Using Constellations}
\titlerunning{Ursa Neural Network}  % abbreviated title (for running head)
%                                     also used for the TOC unless
%                                     \toctitle is used
%

\author{  %Add author(s) here for final submission
  Mark B. Skouson 
  \and Brett J. Borghetti
  \and Robert C. Leishman
}
\institute{
%Add organization(s) here for final submission
  Department of Electrical and Computer Engineering\\
  United States Air Force Institute of Technology\\
  Wright Patterson AFIT, OH 45433
  %\and
  %Department of Electrical and Computer Engineering\\
  %United States Air Force Institute of Technology\\
  %Wright Patterson AFIT, OH 45433
}

\maketitle

\begin{abstract}
This paper describes a neural network layer, named Ursa, that uses a constellation of points to learn classification information from point cloud data. Unlike other machine learning classification problems where the task is to classify an individual high-dimensional observation, in a point-cloud classification problem the goal is to classify a set of d-dimensional observations.  Because a point cloud is a set, there is no ordering to the collection of points in a point-cloud classification problem.  Thus, the challenge of classifying point clouds inputs is in building a classifier which is agnostic to the ordering of the observations, yet preserves the d-dimensional information of each point in the set.  This research presents Ursa, a new layer type for an artificial neural network which achieves these two properties. Similar to new methods for this task, this architecture works directly on d-dimensional points rather than first converting the points to a d-dimensional volume. The Ursa layer is followed by a series of dense layers to classify 2D and 3D objects from point clouds. Experiments on ModelNet40 and MNIST data show classification results comparable with current methods, while reducing the training parameters by over 50 percent. 
\keywords{Point Cloud Classification, Point Sets, 3D Vision, Machine Learning, Deep Learning}
\end{abstract}

\section{Introduction}\label{introsect}
A large bulk of the recent computer vision research has focused on applying artificial neural networks to 2D images. More recently, a growing research area focuses on applying neural networks to 3D physical scenes.  Point clouds or point sets are a common format for representing 3D data since some sensors, including laser-based systems, collect scene data directly as point clouds.  Voxelization is a straightforward way of applying powerful deep convolutional neural network techniques to point clouds, as is done in VoxNet \cite{Maturana2015VoxNet:Recognition} and 3DShapeNets \cite{Wu20153dShapes}.   Voxelization, however, is not always desirable because point clouds can, in many cases, represent structural information more compactly and more accurately than voxelized alternatives. 

In contrast to voxelization methods, PointNet \cite{Qi2017PointNet:Segmentation} and others have developed architectures that operate directly on point clouds, including ECC \cite{Simonovsky2017DynamicGraphs}, Kd-Net \cite{Klokov2017EscapeModels}, DGCNN \cite{Wang2018DynamicClouds}, and KCNet \cite{Shen2018MiningPooling}. This research adds to the growing body of knowledge about learning on point sets.  

This paper describes a neural network layer (Ursa layer) that accepts a point cloud as input and efficiently yields a single feature vector, which is both agnostic to the ordering of the points in the point cloud and encodes the dimensional features of every point.  This output feature vector is an efficient representation of the entire point cloud - an observation which can be used for classification (or other machine learning tasks) in later portions of the network. The layer's trainable parameters are centroids, and each centroid has the same dimension as a point in the point cloud.  For the remainder of this paper, in order to distinguish the centroids from the points in the point cloud, the centroids will be referred to as stars, and the collection of stars in the Ursa layer will be referred to as a constellation. The output of the layer is a feature vector with length equal to the number of constellation stars, which us used in the later layers of the neural network to inform the classification output.  Another important characteristic of this approach is that it does not require a preprocessing step - the Ursa layer is trained as part of the overall network structure using backpropagation and gradient descent learning.

The Ursa layer is invariant to the ordering of the input points. The Ursa layer is not inherently invariant to shift, scale, or rotation; rather, it relies on demonstrations of those types of variations (possibly through data augmentation) in the input data during training to learn these variations.  The output of the Ursa layer is a global shape descriptor of the point cloud that is fed to later layers in the classifier to classify the point cloud. 

Experiments on this architecture show the classification accuracy is comparable to current point cloud-based classifiers, but with a significantly smaller model size. The experiments tested the Ursa architecture with various distance functions and various numbers of constellation stars using MNIST (2D) data and ModelNet40 (3D) data.  Experimentally, the best distance measure was dependent on the data set. For both data sets, too few or too many stars generally degraded performance. Performance gains leveled off with 256 or more stars and, in some cases, more stars led to worse performance.

\section{Selected Related Works}\label{relatedworksect}
The work presented herein is informed by the PointNet \cite{Qi2017PointNet:Segmentation} research and architecture. In \cite{Qi2017PointNet:Segmentation}, Qi, et al., introduce the concept of symmetric functions for unordered points. A symmetric function aggregates the information from each point and outputs a new vector that is invariant to the input order.  Example symmetric operators are summation, multiplication, maximum, and minimum.  Alternatives to a symmetric function for point order invariance would be to sort the input into a canonical order or augment the training data with all kinds of permutations.  PointNet uses a 5-layer multi-layer perceptron (MLP) to convert the input points to a higher-dimensional space, then uses max pooling as the symmetric function to generate a single global feature, which is then fed through 3-layer MLP for classification.  The architecture experimented on in this paper replaces PointNet's first 5 MLP layers and max pooling layer with a single Ursa layer. The Ursa layer generates a global feature and, as in PointNet, uses a 3-layer MLP for classification.  As with PointNet, the Ursa layer's output is invariant to the order of the input data.

This work is also closely related to the KCNet architecture \cite{Shen2018MiningPooling}. KCNet uses a concept similar to the Ursa constellation layer, which they call kernel correlation. Kernel correlation has been used for point set registration, including by \cite{Tsin2004ARegistration}.  Whereas \cite{Tsin2004ARegistration} attempts to find a transformation between two sets of points to align them, Ursa and KCNet allow each point in the constellation (or kernel) to freely move and adjust during training. The KCNet architecture maintains all the layers of PointNet, and augments them by concatenating kernel correlation information to the intermediate vectors within the 5 layers of MLP.   In a forward pass in PointNet, each input point is treated independently of all other points until the global max pooling layer, but that is not the case for KCNet. KCNet uses a set of kernels that operate on local subsets of the input points using a K-nearest neighbor approach. The kernels are trained to learn local feature structures important for classification and segmentation. Thus, KCNet improves on PointNet by adding additional local geometric structure and feature information prior to global max pooling.

There are several difference between the KCNet and the Ursa-based architecture used for this paper.  The KCNet kernel correlation produces a scalar value while the Ursa layer produces a vector. KCNet uses several kernels at the local level to augment the PointNet architecture.  The Ursa architecture uses a single star constellation at the global level to replace the first several layers of PointNet. 

Other deep learning methods that operate on point clouds include Dynamic Graph Convolutional Neural Networks (DGCNNs) \cite{Wang2018DynamicClouds}, Edge-Conditioned Convolution (ECC) \cite{Simonovsky2017DynamicGraphs}, Kd-Networks \cite{Klokov2017EscapeModels}, and OctNet \cite{Riegler2017OctNet:Resolutions}.  These methods organize the data into graphs.  In the cases of \cite{Wang2018DynamicClouds} and \cite{Simonovsky2017DynamicGraphs}, the graphs are based on a vertex for each point and edges that define a relationship between the vertex and near neighbors, and weighted sum operations that operate on vertices and edges of the graph. The DGCNN architecture in \cite{Wang2018DynamicClouds} is quite similar to the PointNet structure, but where the multi-layer perceptron layers are replaced with Edge Convolution Layers. Both \cite{Klokov2017EscapeModels} and \cite{Riegler2017OctNet:Resolutions} use non-uniform spatial structure to partition the input space, and they also used weighted sum operations.  In contrast to these methods, the learning in the Ursa layer is not stored in the weights of a weighted sum operation. Instead the learning is stored in the locations of a set of constellation stars as will be described in the next section. 

This work explores the use of radial basis functions (RBFs) in point cloud classification and so bears some commonality with RBF networks \cite{Buhmann2003RadialImplementations,Orr1996IntroductionNetworks,Broomhead1988MultivariableNetworks,Chen1991OrthogonalNetworks}.  RBFs are able to project an input space into a higher-dimensional space .  It does this through a radial function, which varies  with distance from a central point, and a set of vectors known as RBF centers. In common usage, a function $f(x)$ of an input vector $x$ can be modeled as a weighted sum of a radial basis function, $\phi$, of the distances between the input vector and $m$ RBF centers, $q_i$: 
\begin{equation}
	f(x)  \approx \sum_{i = 1}^{m} \omega_i \left( \phi(||x-q_i||)  \right)
    \label{eq:gau_simple}
\end{equation}
where $|| \cdot ||$ is the L2 norm. 

In general, the $\omega_i$, the $q_i$, and $\phi$ can be selected or trained to fit the RBF network to the function. In practice, $\phi$ and the $q_i$ are usually first selected, then the $\omega_i$ are adjusted or trained to fit the data. While the work herein does explore the use of RBFs, it differs from RBF networks. Rather than computing a weighted sum of RBF outputs to determine a classification of a single input vector (a point cloud), Ursa uses an RBF to transform a set of input vectors to a higher-dimensional feature vector as will be described below.

%In \cite{Alexandridis2017ANetworks}, Alexandridis, et al., describe an RBF network to classify categorical datasets, which is of particular interest to this work because the categorical RBF network receives as input categorical tuples, and point cloud data could be considered a type of categorical tuple data.  For example, each of the 3 dimensions in 3D point cloud data could be considered a category with the value of the category simply be distance from the origin in the associated dimension.  Whereas in \cite{Alexandridis2017ANetworks}, the category values are discrete, in our cases the point cloud data can take on continuous values. 

\section{Method}\label{methodsect}
\begin{figure}
	\centering
		\includegraphics[width=1.00\textwidth]{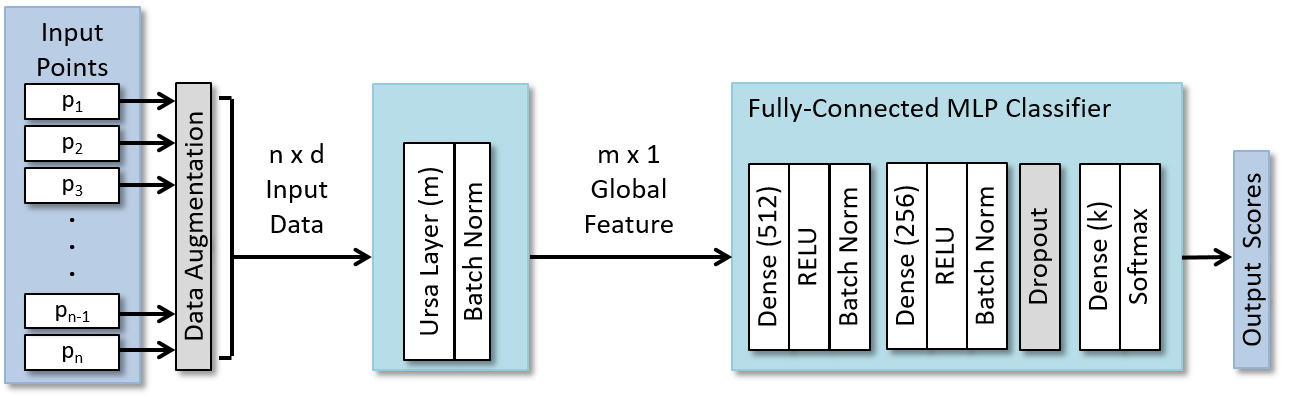}
	\caption{The Ursa architecture. The first hidden layer is an Ursa layer. The remaining three hidden layers are fully-connected (dense) layers. Data augmentation and dropout (the gray boxes) are used only during training.}
	\label{fig:Ursa}
\end{figure}

This section describes the Ursa layer and a neural network architecture that uses a Ursa layer to classify objects.  The overall classification architecture is shown in Fig. \ref{fig:Ursa}.  It is an Ursa layer followed by a three-layer fully connected (dense) multi-layer perceptron classifier.  Compared to the PointNet architecture used by several researchers \cite{Qi2017PointNet:Segmentation,Qi2017PointNet++:Space,Wang2018DynamicClouds,Shen2018MiningPooling}, the Ursa layer replaces the first 5 MLP layers of PointNet architecture, while maintaining the last three-layer MLP portion.  Maintaining an end structure similar to other methods aids in comparison. 

During training, the neural network model makes use of data augmentation at the input and data dropout just prior to the final MLP layer. Because the final layers are straightforward, the remainder of this section will focus on only the the Ursa layer.  Parameters used during implementation are discussed in Section \ref{experimentssect}.

To define the Ursa layer, consider a set of \(n\) \(d\)-dimensional input points in $\Re^d$ that make up a point cloud $P = \{p_1,...,p_n\} \subset \Re^d$. $P$ is the input to the Ursa layer. Within the layer is a constellation of $m$ stars, with the same dimensionality as the input points, $Q = \{q_1,...,q_m\} \subset \Re^d$.  The output of the layer is an $m \times 1$ vector $V = \{v_1,...,v_m\} \subset \Re$.  The Ursa layer converts a set of $n$ $d$-dimensional points into an $m$-dimensional feature vector.  There are $m \times d$ trainable parameters within the layer.

As mentioned earlier, RBFs have the ability to convert vectors to a higher-dimensional feature space. The Ursa layer makes use of RBFs for this purpose. In fact, the experiments investigated the use of three different candidate RBFs and compared their effectiveness.  The first function explored is the Gaussian RBF, $\phi(\cdot) = \exp(-(\cdot)^2/(2 \sigma^2))$, with which the relationship between $P$, $Q$, and $V$ is   
\begin{equation}
	v_i  =\sum_{j = 1}^{n} \exp\left( - \frac{\left\| p_j - q_i\right\|^2}{2\sigma^2}  \right), \hspace{15pt} i= 1,...,m
    \label{eq:gau}
\end{equation}
where $\sigma$ controls the "width" of the function. In this paper, $\sigma$ is considered a user-selected hyper-parameter, potentially tunable in cross-validation.  So, each input point's contribution to the $i$-th entry in the output vector is a function its distance to the $i$-th constellation star, with the contribution decreasing according to the Gaussian function as the point is farther away. In other words, $v_i$ accumulates into a single scalar value the weighted distance information between the $i$th star and each point in the point cloud.  The summation provides the symmetry that makes the output of the layer invariant to the ordering of the input points.

\begin{figure}
	\centering  
		\subfigure[]{\includegraphics[width=0.3\linewidth]{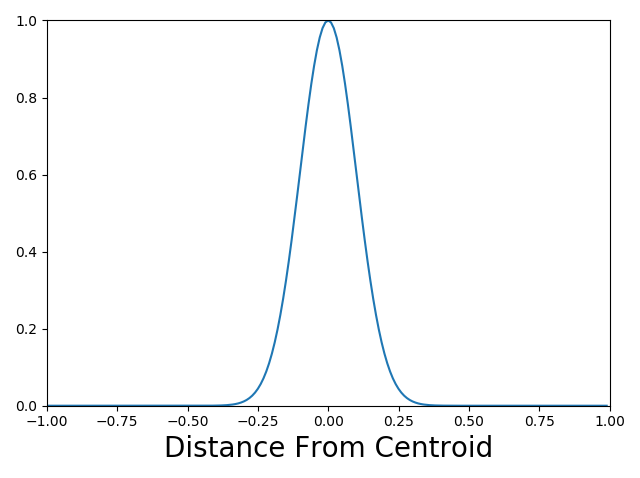}}
		\subfigure[]{\includegraphics[width=0.3\linewidth]{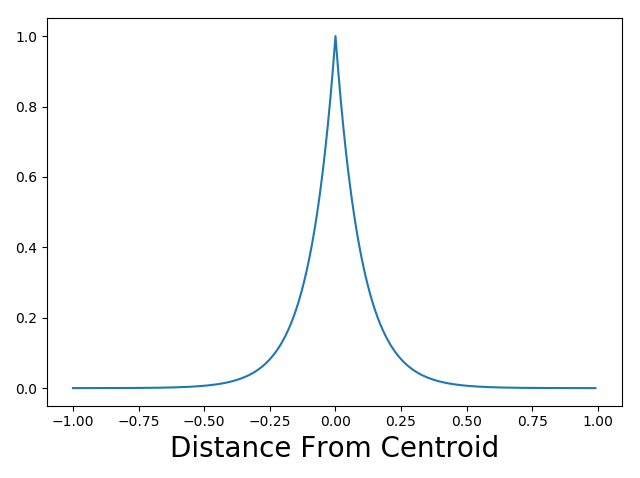}}
		\subfigure[]{\includegraphics[width=0.3\linewidth]{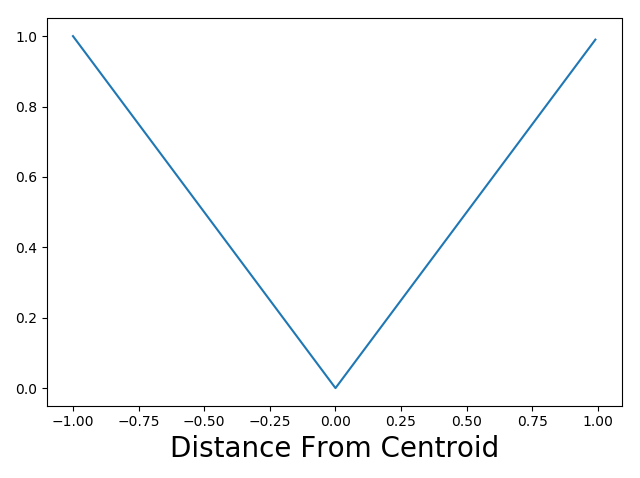}}
	\caption{Visual comparison of RBFs, where the x-axis is the distance from the centroid. (a) Gaussian RBF, (b) exponential decay RBF, (c) RBF for the minimum distance function.}
    \label{fig:gauvsexp}
\end{figure}

The second function explored was an exponential decaying RBF, applied according to Eq. \ref{eq:exp}, where the hyper parameter $\lambda$ controls the width of the function similar to $\sigma$ in Eq. \ref{eq:gau}. 
\begin{equation}
	v_i  =\sum_{j = 1}^{n} \exp \left( - \lambda \left\| p_j - q_i\right\|   \right), \hspace{15pt} i= 1,...,m
    \label{eq:exp}
\end{equation}
The exponential decay has the effect of more rapidly depreciating the contribution of each point to a star's feature output the further they are from the star's location, as can be seen in Figure \ref{fig:gauvsexp}.

 Finally, the experimentation explored a linear RBF applied according to Eq. \ref{eq:minimum}. 
\begin{equation}
	v_i  =\min_{1 \leq j \leq n}\left\| p_j - q_i\right\|, \hspace{15pt} i= 1,...,m
    \label{eq:minimum}
\end{equation}
In this case, the symmetry is provided by the minimum function. The effect of Eq. \ref{eq:minimum} is that $v_i$ is the distance from constellation star $q_i$ to its nearest point from the point cloud. This is the RBF that provides the most efficient computation of the three investigated.  The relative effectiveness of all three measures is shown in Section \ref{experimentssect}.  This paper refers to Eqs. \ref{eq:gau}, \ref{eq:exp}, and \ref{eq:minimum} as the Gaussian distance function, the exponential distance function, and the minimum distance function, respectively, throughout this paper.

The Ursa layer is followed by a three-layer MLP. The non-linearity for each MLP layer is the ReLU function, except for the final layer, which uses softmax. Each ReLU is followed by a batch normalization. The Ursa layer does not require a ReLU function afterward because the $v_i$ in Eqs. \ref{eq:gau}, \ref{eq:exp}, and \ref{eq:minimum} are already always non-negative. Additionally, the Ursa distance measures defined by these equations are not matrix multiplies, which require a separate non-linearity afterward to enable the network to emulate non-linear function of the input; the L2 norm within the computations provides an inherent nonlinear component to the layer. All three distance measures are differentiable and are trained as part of of the overall back-propagation of the entire network. For this implementation, the gradient was computed using the standard tensorflow gradient calculations. 

\section{Experiments}\label{experimentssect}
A series of experiments evaluated the described Ursa network architecture for classification of 3D and 2D objects.  For 3D data, the ModelNet40 shape database \cite{Wu20153dShapes} was selected.  For 2D data, the MNIST handwritten character recognition database \cite{Lecun1998Gradient-basedRecognition} was converted to 2D point clouds and used.

For the ModelNet40 data, 2048 points per object were evenly sampled on mesh faces and normalized into the unit sphere as provided by \cite{Qi2017PointNet++:Space}.  To convert an MNIST image to a 2D point cloud, the coordinates of all pixels with values larger than 128 were used. The maximum number of pixels greater than 128 for any MNIST image was 312.  For those images with fewer than 312 points, the available points from the set were randomly repeated to reach 312 points.  

During training for both 3D and 2D data, the data was augmented by scaling the shape to between 0.8 and 1.25 of the unit-sphere size with a random uniform distribution; rotating the shape between -0.18 and 0.18 radians along each angular axis with a random normal distribution (clipped) with standard deviation 0.06; shifting the shape in every dimension between -0.1 and 0.1 away from its original position with a random uniform distribution; and adding jitter between -0.05 and 0.05 to each point according to a random normal distribution (clipped) with standard deviation 0.01. Also during training a dropout layer was used with a dropout rate of 0.3 just before the last dense layer.  The value for $\sigma$ in Eq. \ref{eq:gau} was chosen to be 0.1, and  $\lambda$ in Eq. \ref{eq:exp} was chosen to be 10.  Additional tuning of these hyper-parameters may improve performance.

The trainable parameters within the Ursa layer were initialized by randomizing them according to a uniform distribution between $\pm$1 in each dimension. The trainable parameters for the dense layers were initialized using the glorot uniform method.  Future research may consider other Ursa layer initialization methods such as uniformly distributing across the space \cite{Broomhead1988MultivariableNetworks} or using information from the input data , e.g. k-means clustering techniques. 

\begin{figure}
	\centering
		\includegraphics[width=1.00\textwidth]{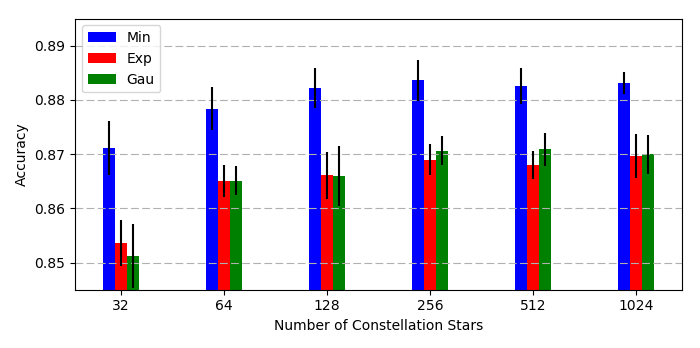}
	\caption{%{\bf ModelNet40 Experimental Results.} 
    The performance of the Ursa architecture on ModelNet40 for each of the distance measures with respect to the number of constellations stars. The y-axis limits have been selected to highlight the small differences. The minimum distance measure slightly outperforms the other two methods.}
	\label{fig:MOdelNet40Graph}
\end{figure}

\begin{figure}
	\centering
		\includegraphics[width=1.00\textwidth]{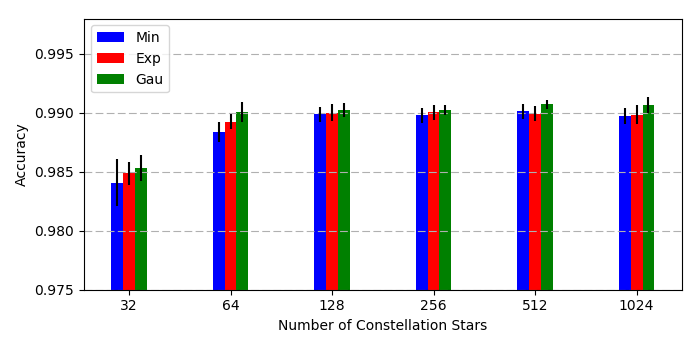}
	\caption{%{\bf MNIST Experimental Results.} 
    The performance of the Ursa architecture on the MNIST data for each of the distance measures with respect to the number of constellations stars. The y-axis limits have been selected to highlight differences. The Gaussian distance measure slightly outperforms the other two methods.}
	\label{fig:MNISTGraph}
\end{figure}

Experiments explored the three feature space transforms in Eqs. \ref{eq:gau}, \ref{eq:exp}, and \ref{eq:minimum} for several values of $m$, the number of constellation stars.  Classification performance was evaluated for $m=$ 32, 64, 128, 256, 512, and 1024.  Ten independent tests were conducted with each of the distance measures at each value of $m$.  The average accuracy is plotted in Figs. \ref{fig:MOdelNet40Graph} and \ref{fig:MNISTGraph}.  


\section{Results and Discussion}
An analysis of Figs. \ref{fig:MOdelNet40Graph} and \ref{fig:MNISTGraph} shows a general trend of performance improving steadily as the constellation grows to 256 stars, but leveling out or perhaps worsening beyond 256 stars. It is interesting that the experiments did not show any clear difference in the number of stars needed based on the dimensionality of the data (3D vs 2D), number of points per shape (2048 vs 312), or number of possible classes (40 vs 10). An in-depth analysis revealed that when the number of stars increased over 512, many of the constellation stars were effectively unused; they were pushed to the edges of the 2- or 3-dimensional space. The data suggest a good starting point for other data sets may be 256 or 512 stars.

The minimum distance function performed slightly better on the ModelNet40 data, while the Gaussian distance function performed slightly better on the MNIST data. This may be a function of the number of points per shape.  The minimum function may be more effective in situations with many points. It is interesting to note that PointNet \cite{Qi2017PointNet:Segmentation} performed better using max pooling rather average pooling or an attention weighted sum to provide the symmetry (point order invariance) property. The minimum distance measure has similarity to a max pooling, while the Gaussian distance function uses a weighted sum.

Table \ref{table:compareresults} shows how the evaluated Ursa network compares to other classification methods on the same data sets. Ursa achieved impressive results, especially considering the Ursa model uses far fewer parameters than any other method compared.  While some more sophisticated methods outperformed Ursa, this paper demonstrates the viability, effectiveness, and potential of the Ursa layer and the constellation approach to pattern learning.

\begin{table}
  \caption{Classification results and model size comparisons for various methods. Accuracy results for the Ursa method are the mean of 10 independent runs for each data set.  Best Ursa single run performances are in parentheses. Accuracy results for the other methods are adapted from \cite{Shen2018MiningPooling} and \cite{Wang2018DynamicClouds}. Model size for Ursa is based on 512 stars.}
  \centering
  \begin{tabular}{lccc}
    \toprule
      & ModelNet40 Accuracy  & MNIST Accuracy  & Model Size  \\
     &  (in percent) &  (in percent) &  (x1M params) \\
    \midrule
    LeNet5 \cite{Lecun1998Gradient-basedRecognition} & -- & 99.2 & --  \\
    3DShapeNets \cite{Wu20153dShapes} & 84.7 & -- & --   \\
    VoxNet \cite{Maturana2015VoxNet:Recognition} & 85.9 & -- & --   \\
    Subvolume \cite{Qi2016VolumetricData} & 89.2  & -- & --  \\
    ECC \cite{Wu20153dShapes} & 87.4 & 99.4 & --  \\
    PointNet (Baseline) \cite{Qi2017PointNet:Segmentation} & 87.2 & 98.7 & 0.8 \\
    PointNet \cite{Qi2017PointNet:Segmentation} & 89.2 & 99.2 & 3.5  \\
    PointNet++ \cite{Qi2017PointNet++:Space} & 90.7 & 99.5 & 1.0   \\
    KCNet \cite{Shen2018MiningPooling} & 90.0 & 99.3 & 0.9    \\
    Kd-Net \cite{Klokov2017EscapeModels} & 91.8 & 99.1 &  2.0 \\
    DGCNN \cite{Wang2018DynamicClouds} & 92.2 & -- & 1.8  \\
	\midrule
    Ursa (Ours) & 88.2 (89.0) & 99.1 (99.2) & 0.4    \\
    \bottomrule
  \end{tabular}
  \label{table:compareresults}
\end{table}

\begin{figure}
	\centering  
		\subfigure[]{\includegraphics[width=0.3\linewidth]{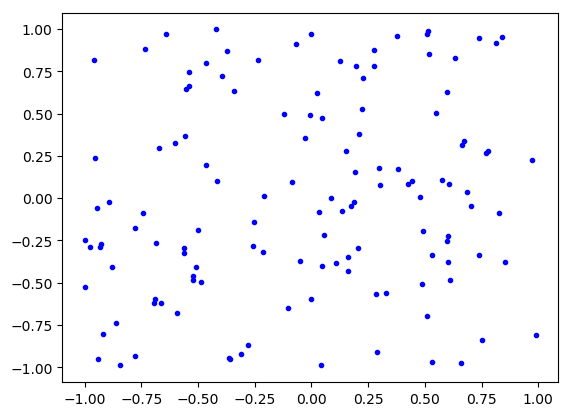}}
		\subfigure[]{\includegraphics[width=0.3\linewidth]{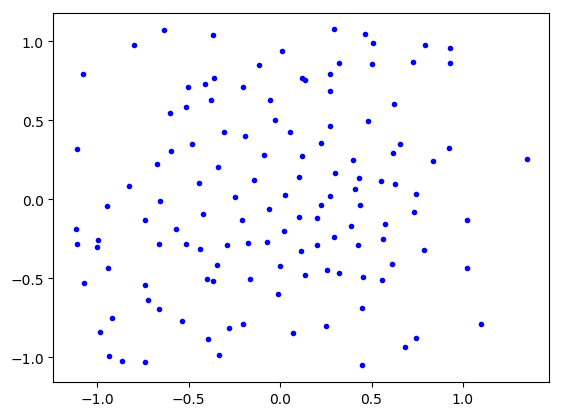}}
		\subfigure[]{\includegraphics[width=0.3\linewidth]{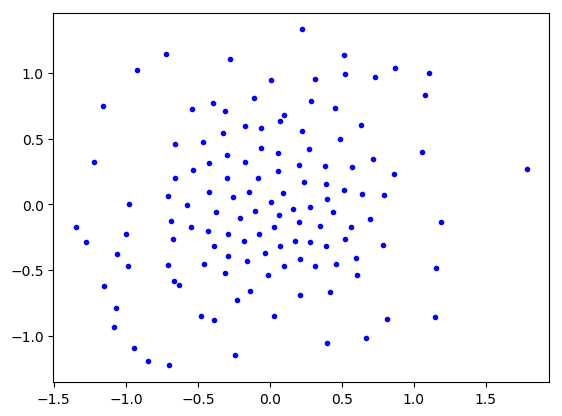}}
		\subfigure[]{\includegraphics[width=0.3\linewidth]{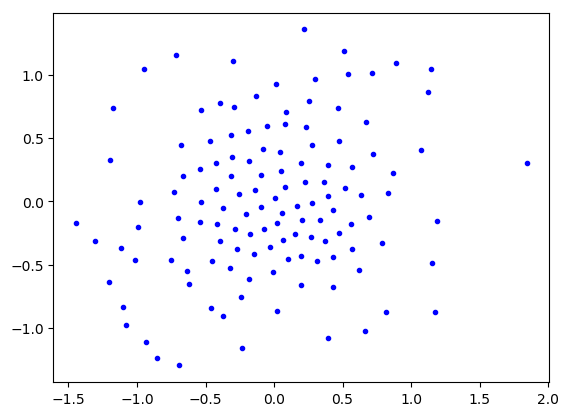}}
		\subfigure[]{\includegraphics[width=0.3\linewidth]{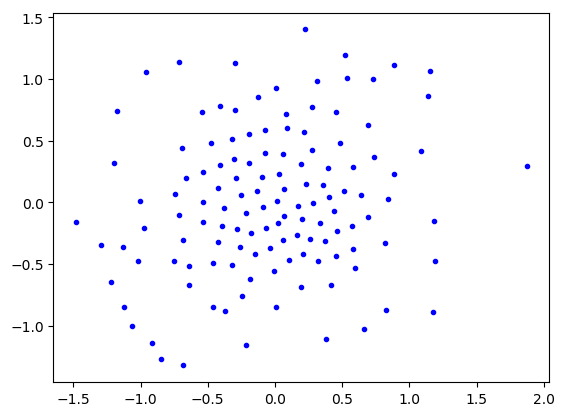}}
		\subfigure[]{\includegraphics[width=0.3\linewidth]{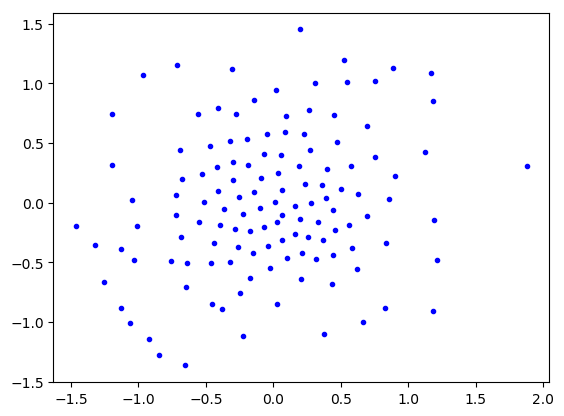}}
	\caption{Depiction of the Ursa constellation stars and their positions during training for for m=128 and using the minimum distance measure. Sub-figures show (a) random initialization, (b) after 10 training epochs, (c) after 100 epochs, (d) after 200 epochs, (e) after 300 epochs, and (f) after 500 epochs.}
    \label{fig:minweights}
\end{figure}

\begin{figure}
	\centering  
		\subfigure[]{\includegraphics[width=0.3\linewidth]{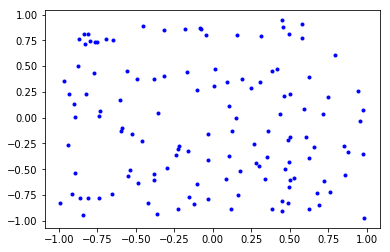}}
		\subfigure[]{\includegraphics[width=0.3\linewidth]{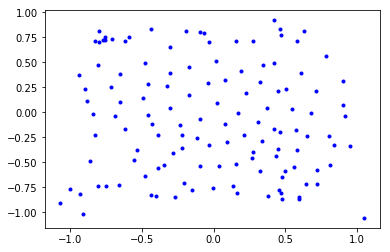}}
		\subfigure[]{\includegraphics[width=0.3\linewidth]{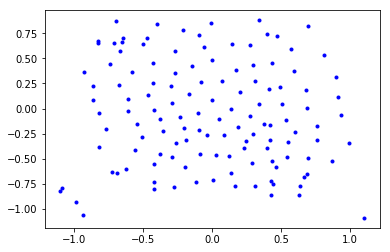}}
		\subfigure[]{\includegraphics[width=0.3\linewidth]{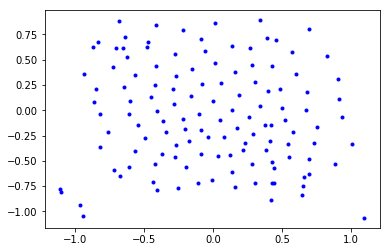}}
		\subfigure[]{\includegraphics[width=0.3\linewidth]{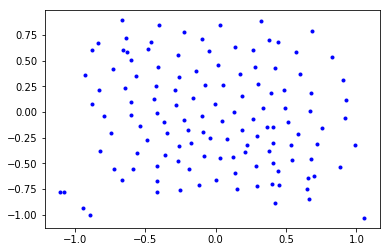}}
		\subfigure[]{\includegraphics[width=0.3\linewidth]{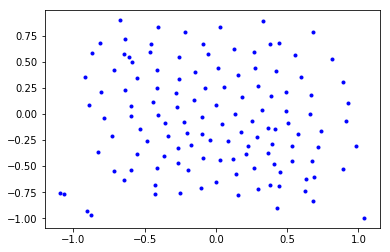}}
	\caption{Depiction of the Ursa constellation stars and their positions during training for for m=128 and using the Gaussian distance measure. Sub-figures show (a) random initialization, (b) after 10 training epochs, (c) after 100 epochs, (d) after 200 epochs, (e) after 300 epochs, and (f) after 500 epochs.  }
    \label{fig:rbfweights}
\end{figure}

The 2D data was used to demonstrate the movement over time of the Ursa constellation stars during training.  Figs. \ref{fig:minweights} and \ref{fig:rbfweights} show the constellation stars adjusting over time during training to span the space for the minimum distance measure of Eq. \ref{eq:minimum} and the Gaussian distance measure of Eq. \ref{eq:gau}, respectively.  The constellation resulting from the minimum distance measure appears more compact in the center and the stars are more spread out toward the outside.  On the other hand, the Gaussian distance measure constellation is more uniformly distributed throughout the space. Also, the resulting range is larger for the minimum distance measure than for the Gaussian distance measure. 

As mentioned in Section \ref{relatedworksect}, The Ursa layer computes a single global feature for a set of points, so it is limited in its ability to recognize local structures and patterns.  Future research should explore a hierarchical network architecture based on the Ursa layer that evaluates object structure at varying levels. The current architecture is also not invariant to shifts, scales, and rotations of the input data. This is another area for future research.

\section{Conclusion}\label{conclusionsect}
This paper has presented an Ursa neural network layer and demonstrated its effectiveness and viability for classification of point cloud data. The Ursa layer stores information in the form of constellation points, rather than a set of multiplicative weights in a matrix. While other more sophisticated methods achieved higher classification rates on the data sets used in testing, all other methods compared used at least twice the model parameters of the baseline Ursa network. 

\bibliographystyle{ieeetr}
\bibliography{Mendeley}

\begin{thebibliography}{10}

\bibitem{Maturana2015VoxNet:Recognition}
D.~Maturana and S.~Scherer, ``{VoxNet: A 3D Convolutional Neural Network for
  real-time object recognition},'' in {\em 2015 IEEE/RSJ International
  Conference on Intelligent Robots and Systems (IROS)}, pp.~922--928, IEEE, 9
  2015.

\bibitem{Wu20153dShapes}
Z.~Wu, ``{3d shapenets: A deep representation for volumetric shapes},'' in {\em
  IEEE Conference on Computer Vision and Pattern Recognition, CVPR 2015},
  p.~1912–1920, 2015.

\bibitem{Qi2017PointNet:Segmentation}
C.~R. Qi, H.~Su, K.~Mo, and L.~J. Guibas, ``{PointNet: Deep learning on point
  sets for 3D classification and segmentation},'' in {\em Proceedings - 30th
  IEEE Conference on Computer Vision and Pattern Recognition, CVPR 2017},
  vol.~2017-Janua, pp.~77--85, 2017.

\bibitem{Simonovsky2017DynamicGraphs}
M.~Simonovsky and N.~Komodakis, ``{Dynamic edge-conditioned filters in
  convolutional neural networks on graphs},'' in {\em Proceedings - 30th IEEE
  Conference on Computer Vision and Pattern Recognition, CVPR 2017},
  pp.~29--38, 2017.

\bibitem{Klokov2017EscapeModels}
R.~Klokov and V.~Lempitsky, ``{Escape from Cells: Deep Kd-Networks for the
  Recognition of 3D Point Cloud Models},'' {\em Proceedings of the IEEE
  International Conference on Computer Vision}, vol.~2017-Octob, pp.~863--872,
  2017.

\bibitem{Wang2018DynamicClouds}
Y.~Wang, Y.~Sun, Z.~Liu, S.~E. Sarma, M.~M. Bronstein, and J.~M. Solomon,
  ``{Dynamic Graph CNN for Learning on Point Clouds},'' in {\em
  arXiv:1801.07829}, 2018.

\bibitem{Shen2018MiningPooling}
Y.~Shen, C.~Feng, Y.~Yang, and D.~Tian, ``{Mining Point Cloud Local Structures
  by Kernel Correlation and Graph Pooling},'' in {\em IEEE Conference on
  Computer Vision and Pattern Recognition}, 2018.

\bibitem{Tsin2004ARegistration}
Y.~Tsin and T.~Kanade, ``{A Correlation-Based Approach to Robust Point Set
  Registration},'' in {\em European Conference on Computer Vision}, 2004.

\bibitem{Riegler2017OctNet:Resolutions}
G.~Riegler, A.~O. Ulusoy, and A.~Geiger, ``{OctNet: Learning deep 3D
  representations at high resolutions},'' in {\em Proceedings - 30th IEEE
  Conference on Computer Vision and Pattern Recognition, CVPR 2017},
  vol.~2017-Janua, pp.~6620--6629, 2017.

\bibitem{Buhmann2003RadialImplementations}
M.~D. Buhmann, {\em {Radial basis functions : theory and implementations}}.
\newblock Cambridge University Press, 2003.

\bibitem{Orr1996IntroductionNetworks}
M.~Orr, {\em {Introduction to Radial Basis Function Networks}}.
\newblock 1996.

\bibitem{Broomhead1988MultivariableNetworks}
D.~S. Broomhead and D.~Lowe, ``{Multivariable Functional Interpolation and
  Adaptive Networks},'' {\em Complex Systems}, vol.~2, pp.~3221--355, 1988.

\bibitem{Chen1991OrthogonalNetworks}
S.~Chen, C.~Cowan, and P.~Grant, ``{Orthogonal least squares learning algorithm
  for radial basis function networks},'' {\em IEEE Transactions on Neural
  Networks}, vol.~2, pp.~302--309, 3 1991.

\bibitem{Qi2017PointNet++:Space}
C.~R. Qi, L.~Yi, H.~Su, and L.~J. Guibas, ``{PointNet++: Deep Hierarchical
  Feature Learning on Point Sets in a Metric Space},'' in {\em Neural
  Information Processing Systems Conference}, 2017.

\bibitem{Lecun1998Gradient-basedRecognition}
Y.~Lecun, L.~Bottou, Y.~Bengio, and P.~Haffner, ``{Gradient-based learning
  applied to document recognition},'' {\em Proceedings of the IEEE}, vol.~86,
  no.~11, pp.~2278--2324, 1998.

\bibitem{Qi2016VolumetricData}
C.~R. Qi, H.~Su, M.~Niessner, A.~Dai, M.~Yan, and L.~J. Guibas, ``{Volumetric
  and Multi-View CNNs for Object Classification on 3D Data},'' {\em Proceedings
  - 29th IEEE Conference on Computer Vision and Pattern Recognition, CVPR
  2016}, 2016.

\end{thebibliography}

\end{document}